\documentclass[twoside,11pt]{article}

\newlength{\defaultfigsize}
\usepackage{etoolbox}
\AtBeginEnvironment{tabular}{\scriptsize}

\usepackage{jmlr2e}

\usepackage{multirow}
\usepackage{microtype}
\usepackage{graphicx}
\usepackage{subfigure}
\usepackage{booktabs} 
\usepackage{amsmath}
\usepackage{amsfonts}
\usepackage{algorithm, algorithmic}
\usepackage{array}
\ShortHeadings{Disease-Atlas: Navigating Disease Trajectories using Deep Learning}{Bryan Lim and Mihaela van der Schaar}
\firstpageno{1}

\begin{document}

\title{Disease-Atlas: Navigating Disease Trajectories using Deep Learning}
\author{\name Bryan Lim \email bryan.lim@eng.ox.ac.uk \\
       \addr Department of Engineering Science\\ 
       University of Oxford,
       Oxford, UK
       \AND
       \name Mihaela van der Schaar \email mihaela.vanderschaar@eng.ox.ac.uk \\
       \addr Department of Engineering Science\\
       University of Oxford,
       Oxford, UK\\
       }
%
\maketitle

\begin{abstract}
Joint models for longitudinal and time-to-event data are commonly used in longitudinal studies to forecast disease trajectories over time. While there are many advantages to joint modeling, the standard forms suffer from limitations that arise from a fixed model specification and computational difficulties when applied to high-dimensional datasets. In this paper, we propose a deep learning approach to address these limitations, enhancing existing methods with the inherent flexibility and scalability of deep neural networks while retaining the benefits of joint modeling. Using longitudinal data from two real-world medical datasets,  we demonstrate improvements in performance and scalability, as well as robustness in the presence of irregularly sampled data.
\end{abstract}

\section{Introduction}

Building a Disease Atlas for clinicians involves the dynamic forecasting of medical conditions based on clinically relevant variables collected over time, and guiding them in charting a course of action. This includes the simultaneous prediction of survival probabilities, risks of developing related diseases, and relevant biomarker trajectories at different stages of disease progression. While prognosis, i.e. survival prediction, is usually the main area of focus \citep{vanHouwelingenDynamicPrediction, RizopoulosDynamicPrediction2017}, a growing area in precision medicine is the forecasting of personalized disease trajectories, using patterns in temporal correlations and associations between related diseases to predict their evolution over time. \citep{TemporalDiseaseTrajectoriesNature, DiseaseTrajectories2017}. Dynamic prediction methods that account for these interactions are particularly relevant in multimorbidity management, as patients with one chronic disease typically develop other long-term conditions over time \citep{Farmeri4843}. With the mounting evidence on the prevalence of multimorbidity in aging populations around the world \citep{MultimorbidityEvidence}, the ability to jointly forecast multiple clinical variables would be beneficial in providing clinicians with a fuller picture of a patient's medical condition.

A substantial portion of machine learning literature investigates predictions with time-series data, typically focusing on patients in the hospital. In this setting, patients are tracked for a relatively short period of time, spanning from a few days to weeks, with measurements collected every few hours. This leads to the collection of numerous measurements, potentially with a high degree of missingness. Given the length of the monitoring period, in-hospital predictions are usually narrow in their scope, focusing on detecting the rapid onset of critical events, such as ICU admission, and not considering the prediction of comorbidities which can take years to develop.

With chronic diseases however, such as cystic fibrosis or diabetes, patients are followed up over the span of years, usually as part of regular physical examinations. This differs significantly from the in-hospital setting as measurements are collected infrequently, e.g. once every few years and possibly at irregular intervals, leading to relatively few observations per patient. The state of the patient also evolves slowly, allowing for the development of related comorbidities over time. Additional comorbidities in turn affect key biomarkers which reflect a patient's clinical state and rate of deterioration, such as lung function scores (e.g. FEV1) in cystic fibrosis or brain scan measurements in Alzheimer's disease. As such, the ability to jointly forecast comorbidity and biomarker trajectories, in addition to survival, allows for early intervention by clinicians to prevent the development of other related diseases and forestall further deterioration. This would allow for an improved quality of life for the patient even if immediate improvements to survival might be small.  Hence, the development of new machine learning methods to combine longitudinal predictions with dynamic survival (time-to-event) analysis, where events-of-interest can include heart failure, respiratory failure or the onset of dementia in addition to death, would allow for a more holistic management of long-term conditions, going above and beyond the short-term survival prediction usually seen in hospital settings.

Traditionally, joint models for longitudinal and time-to-event data have been commonly used in clinical studies when there is prior knowledge indicating an association between longitudinal trajectories and survival. Using individual models for each data trajectory as building blocks, such as linear mixed models for longitudinal data and the Cox proportional hazard model for survival, joint models add a common association structure on top of them, e.g. through shared-random effects or frailty models \citep{Hickey2016JointModelling}. From a dynamic prediction perspective, joint models have been shown to lead to a reduced bias in estimation \citep{BasicConceptsJointModels} and improved predictive accuracy \citep{HoganAndLairdIncreasingEfficiencyJointModels}. However, standard joint models face severe computational challenges when applied to large datasets, which arise when increasing the dimensionality of the random effects component \citep{Hickey2016JointModelling}. 

\begin{figure}[bt]
\centerline{\includegraphics[width=1.75\defaultfigsize]{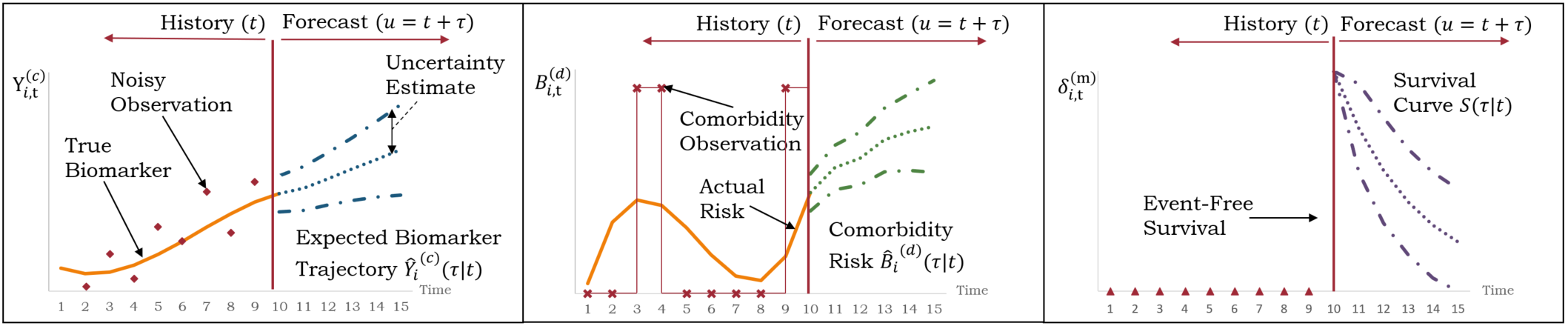}}
\caption{Illustration of Disease-Atlas Predictions over Time}
\label{fig:DynamicPrediction}
\end{figure}
To overcome the limitations of joint models, we introduce Disease-Atlas - a scalable deep learning approach to forecasting disease trajectories over time. Our main contributions are as follows:

\paragraph{Deep Learning for Joint Models} We provide a novel conception of the joint modeling framework using deep learning, capturing the relationships between trajectories through shared representations learned directly from data, and improving scalability as a whole. The network outputs parameters of predictive distributions for longitudinal and time-to-event data that take a similar form to the sub-models used in joint modeling. To the best of our knowledge, this paper is the first to investigate the use of deep learning in joint models for longitudinal and survival trajectories.
\paragraph{Robustness to Irregular Sampling via Multitask Learning} Observations in longitudinal studies are very rarely aligned at every time step, as measurements can be collected at different sampling frequencies. Hence, training a multioutput neural network would require the imputation of the target labels as a pre-processing step, so as to artificially align the dataset prior to calibration. This could lead to poorer predictions if imputation quality is low and a high degree of missingess is present, as the network is biased to simply learn the imputation mechanism. To mitigate this issue, we formulate joint model calibration as a multitask learning problem, grouping variables - which are measured at the same time and with similar sampling frequencies - together into tasks, and training the network using only actual observations as target labels.
\paragraph{Incorporating Medical History into Forecasts} While deep learning for medicine has gained popularity in recent times, the majority of methods, such as \citep{DeepMultitaskGPCompetingRisks, DeepSurvivalAnalysis}, only use covariates at a single time point in making predictions. However, a patient's medical history could also be informative of her future clinical outcomes, and predictions could be improved by incorporating past information. We integrate historical information into our network using a Recurrent Neural Network (RNN) in the base layer, which contains a memory state that updates over time as new observations come in.

\section{Related Work}
\label{sec:RelatedWorks}
While the utility of joint models has been demonstrated by its popularity in longitudinal studies, numerous modeling choices exist, each containing its own advantages and limitations (see \citep{Hickey2016JointModelling} for a full overview). \citep{RizopoulosBayesianModelAveraging}, for example, highlight the sensitivity of predictions to the association structures used, adopting a Bayesian model averaging approach instead to aggregate the outputs of different models over time. In this respect, the flexibility of deep learning has the potential to enhance dynamic predictions with joint models, by directly learning variable relationships from the data itself, and completely removing the need for explicit model specification. In addition, \citep{Hickey2016JointModelling, BarrettJointModellingEfficiency, BoostingJointModels, FutomaScalableJointModelling} note performance limitations when applying standard joint models to high-dimensional datasets. These are typically estimated using Expectation Maximization (EM) or Markov Chain Monte Carlo (MCMC) sampling methods, which rapidly grow in complexity with the number of covariates and random effects. As such, most studies and software packages often focus on modeling a single or a small number of longitudinal measurements, along with a time-to-event of interest. However, the increase in data availability through electronic health records opens up the possibility of using information from multiple trajectories to improve predictions. Recent works have attempted to address this limitation by exploiting special properties of the longitudinal sub-models, such as the multivariate skew-normal structure in \citep{JointModelsBarrett}, and the combination of variational approximation and dynamic EM-style updates over time in \citep{FutomaScalableJointModelling}. In light of this, the use of deep learning holds much promise in enhancing the performance of joint models, given its inherent ability to scale with large datasets without the need for specific modeling assumptions.

Deep learning has seen increasing use in medical applications, with successes in traditional survival analysis \citep{DeepSurvivalAnalysis, DeepLearningSurvival} survival analysis with competing risks \citep{DeepHit,DeepMultitaskGPCompetingRisks} and treatment recommendations \citep{Katzman2016DeepSurvPT}. In general, these methods focus purely on forecasting survival, do not consider dynamic prediction over time and only use covariates at a single time point in making predictions. Deep Kalman Filters \citep{DeepKalmanFilters} use a network which does dynamically update its latent states over time but assumes that all outputs follow the same distribution. This prevents it from being applied to heterogeneous datasets, which limits its usage for joint modeling. 

RNNs with multitask learning have also been used extensively in making predictions inside the hospital, using frequently sampled measurements as inputs for event detection or automated diagnosis (\citep{MultitaskLSTMs, DiseaseOnsetPrediction, ADMultitaskClassification, LearningToDiagnose}). In addition to the characteristics of in-hospital data (i.e. sampled at hourly intervals with a high rate of missing data), these works bear several fundamental differences to the Disease-Atlas. Firstly, in-hospital predictions focus on separate dynamic classification tasks which produce single class labels at each time step, such as the most likely event or diagnosis at the next time step. In contrast, the Disease-Atlas allows for the simultaneous prediction of multiple variables of interest at each time step, which can be either discrete (classification) or continuous (regression). As predictions are conditioned on the same latent structure, we can analyze variables in a consistent fashion and use trajectory forecasts to understand changes in survival probabilities. For example, poorer survival odds could be a result of increased risk of multiple infections. Secondly, the RNNs in existing works typically produce single point forecasts, which do not account for the uncertainty of the model. We address this limitation using the Monte-Carlo Dropout procedure of \citep{GalDropoutRNN} (see Section \ref{sec:dropout}), producing predictions with uncertainty estimates at each time step.

\section{Problem Definition}
For a given longitudinal study, let there be $N$ patients with observations made at time $t$, for $ 0 \le t \le T_{cens}$ where $T_{cens}$ denotes an administrative censoring time \footnote{\label{administrativecensoring} Administrative censoring refers to the right-censoring that occurs when a study observation period ends. }. For the $i^{th}$ patient at time $t$, observations are made for a $K$-dimensional vector of longitudinal variables $\mathbf{V_{i,t}} = [Y_{i,t}^{(1)}, \dots, Y_{i,t}^{(C)}, B_{i,t}^{(1)}, \dots, B_{i,t}^{(D)} ]$, where $Y_{i,t}^{(c)}$ and $B_{i,t}^{(d)}$ are continuous and discrete longitudinal measurements respectively, a $L$-dimensional vector of external covariates $\mathbf{X_{i,t}} = [X_{i,t}^{(1)}, \dots X_{i,t}^{(L)} ]$, and a $M$-dimensional vector of event occurrences $\mathbf{\delta_{i,t}} = [\delta_{i,t}^{(1)}, \dots, \delta_{i,t}^{(M)} ]$ , where $\delta_{i,t}^{(m)} \in \{0, 1\}$ is an indicator variable denoting the presence or absence of the $m^{th}$ event. $T_{i,t}^{(m)}$ is defined to be the first time the event is observed after $t$, which allows us to model both repeated events and events that lead to censoring (e.g. death). The final observation for patient i occurs at $T_{i,\max} =  \min( T_{cens}, T_{i,0}^{(a_1)}, \dots , T_{i,0}^{(a_{\max})})$, where $ \{ a_i, \dots, a_{\max} \} $ is the set of indices for events that censor observations. Furthermore, we introduce a filtration $\mathcal{F}_{i,t}$ to capture the full history of longitudinal variables, external covariates and event occurrences of patient $i$ until time $t$.

\subsection{Joint Modeling}
\label{sec:JointModels}
From \citep{Hickey2016JointModelling}, numerous sub-models for longitudinal measurements exist, each with their own pros and cons. General forms for continuous and binary longitudinal measurements are typically expressed as:
\begin{equation}
\begin{split}
& Y_{i,u}^{(c)} | \mathcal{F}_{i,t}  
\sim  \text{N} \left( m^{(c)}\left(u, \mathcal{F}_{i,t}; \mathbf{b}_{i} , \tilde{ \mathbf{W}}\right), \sigma_u^{(c) ~2}  \right) 
\end{split}
\end{equation}
\begin{equation}
\begin{split}
& B_{i,u}^{(d)} | \mathcal{F}_{i,t} \sim \text{Bernoulli} \left( \Phi^{(d)}\left(u, \mathcal{F}_{i,t}; \mathbf{b}_{i} , \tilde{\mathbf{W}} \right) \right) 
\end{split}
\end{equation}
Where $ m^{(c)}(.)$ is a function for the predictive mean of the $c$-th  longitudinal variable, and $\sigma_t^{(c)~2}$ its variance. $\Phi^{(d)}(.)$ is a function for the probability of the binary observation, such as the commonly used logit or probit functions, and $\tilde{\mathbf{W}}$ is the vector of static coefficients used by the sub-models.

In both models, $\mathbf{b}_{i}$ is a vector of association parameters used across trajectories, and define the association structure of the joint model. While the majority of models use subject-specific random effects, this can also refer to time-dependent latent variables as seen in \citep{latentVariablesLongitudinalModels} or shared spline coefficients in \citep{JointModelsBarrett}. 

Event times can be expressed using the general form below:
\begin{equation}
\begin{split}
& T_{i,t}^{(m)} | \mathcal{F}_{i,t} \sim \mathcal{S} \left( \Lambda^{(m)}\left(t, \mathcal{F}_{i,t}; \mathbf{b}_{i} ,\tilde{\mathbf{W}} \right) \right) 
\end{split}
\end{equation}

Where $\mathcal{S}$ is an appropriate survival distribution (e.g. Exponential, Weibull, etc), and $\Lambda^{(m)}(.)$ is a generic cumulative hazard function. In most joint model applications, this typically takes the form of the Cox proportional hazards model.

The standard linear mixed effects models can be expressed as:
\begin{equation}
\begin{split}
Y_{i,t}^{(c)} &=  \mathbf{X}_{i,t}^\top ~ \mathbf{\theta}_{fix}^{(c)} + \mathbf{R}_{i,t}^\top ~ \mathbf{\theta}^{(c)}_{rand} + \epsilon	^{(c)}_{i,t}  \\
& = m^{(c)} (t) + \epsilon^{(c)}_t
\end{split}
\end{equation}
\begin{equation}
h_{i,t}^{(m)} = h_0(t) \exp\left( \mathbf{B}_{i}^\top \mathbf{\gamma}^{(m)} + \beta^{(m)} m^{(c)} (t)\right)
\end{equation}

Where $\mathbf{X}_{i,t}$ and $\mathbf{R}_{i,t}$ are time-dependent design vectors for fixed effects $\mathbf{\theta}_{fix}^{(c)}$ and random effects $\mathbf{\theta}^{(c)}_{rand}$, $\epsilon_{i,t}^{(c)} \sim N\left(0,\sigma_t ^{(c)~2}\right) $ is a random noise term, and $h_{i,t}^{(m)} $ is the hazard rate of the survival process, with patient fixed covariates $\mathbf{B}_i$, and static coefficients $\gamma^{(m)}$ and $ \beta^{(m)}$.

In this model, we define the association parameters of the joint model to be those common to both longitudinal and survival processes, i.e $\mathbf{b}_{i} = [\mathbf{\theta}_{fix}^{(c)}$, $\mathbf{\theta}^{(c)}_{rand}]$, and static coefficients which are unique to the separate processes, i.e. $\mathbf{\tilde{W}} = [\gamma^{(m)}, \beta^{(m)}]$.


\subsection{Dynamic Prediction}
\label{sec:DynamicPredictionDefinition}
Dynamic prediction in joint models can be defined as the estimation of both the expected values of longitudinal variables and survival probabilities over a specific time window $\tau$ in the future:
\begin{equation}
\label{eqn:DynPredExpectation}
 \hat{V}_i^{(k)}(\tau| t) = \mathbb{E} \left[V_{i,t+\tau}^{(k)}|  \mathcal{F}_{i,t}; \mathbf{b}_{i}, \tilde{\mathbf{W}}  \right]
\end{equation}
\begin{equation}
\label{eqn:survivaldefn}
\begin{split}
 &S_i^{(m)}(\tau| t)
  = P\left(T_{i,0}^{(m)} \ge t+\tau | T_{i,0}^{(m)} \ge t,  \mathcal{F}_{i,t}; \mathbf{b}_{i},  \tilde{\mathbf{W}} \right) 
 \end{split}
\end{equation}
Where $ V_{i,t}^{(k)}$ is the $k^{th}$ longitudinal variable at time t that can be either continuous or binary.
A conceptual illustration of dynamic prediction can be found in Figure \ref{fig:DynamicPrediction}. For continuous longitudinal variables, such as biomarker predictions, the goal is to forecast its expected value given all the information until the current time step (i.e. $\mathcal{F}_{i,t}$). In the case of binary observations, such as the presence of a comorbidity, the expectation in Equation \ref{eqn:DynPredExpectation} is the probability (or risk) of developing a comorbidity at time $t+\tau$. The survival curves shown give us the probability of not experiencing an event over various horizons $\tau$ given $\mathcal{F}_{i,t}$. While the uncertainty estimates are model dependent, these are usually expressed via confidence intervals in frequentist methods, or using the posterior distributions for $\mathbf{b}_{i,t}$ and $\tilde{\mathbf{W}}$ in Bayesian models.
\section{Network Design}
\subsection{Architecture}
\label{sec:Architecture}
\begin{figure*}[bt]
\centerline{\includegraphics[width=1.75\defaultfigsize]{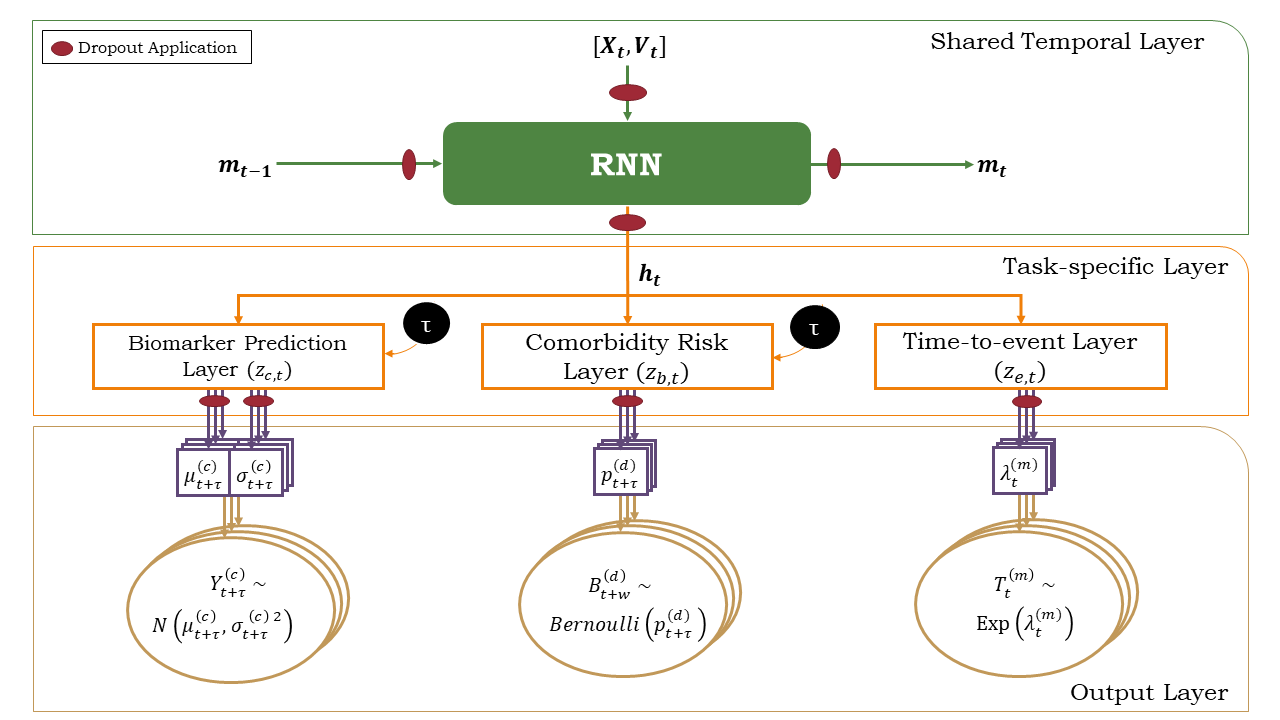}}
\caption{Disease-Atlas Network Architecture}
\label{fig:NetworkArchitecture}
\end{figure*}
Disease-Atlas captures the associations within the joint modeling framework, by learning \textit{shared representations} between trajectories at different stages of the network, while retaining the same sub-model distributions captured by joint models. The network, as shown in Figure \ref{fig:NetworkArchitecture}, is conceptually divided into 3 sections: 1) A shared temporal layer to learn the temporal and cross correlations between variables, 2) task-specific layers to learn shared representations between related trajectories, and 3) an output layer which computes parameters for predictive sub-model distributions for use in likelihood loss computations during training and generating predictive distributions at run-time.

The equations for each layer are listed in detail below. For notational convenience, we drop the subscript $i$ for variables in this section, noting that the network is only applied to trajectories from one patient at time.

\paragraph{Shared Temporal Layer}
We start with an RNN at the base of the network, which incorporates historical information (i.e. $\mathcal{F}_{t}$) into forecasts by updating its memory state over time. For the tests in Section \ref{sec:Experiment}, the usage of both the Simple Recurrent Network (SRN) and LSTM in this layer was compared. 
\begin{align}
[ \mathbf{h_t}, \mathbf{m_t} ] = \text{RNN}(  [\mathbf{X_{t}}, \mathbf{V_{t}}] , \mathbf{m_{t-1}})&
\end{align}
Where $\mathbf{h_t}$ is the output of the RNN and $\mathbf{m_t}$ its memory state. To generate uncertainty estimates for forecasts and retain consistency with joint models, we adopt the MC dropout approach described in \citep{GalDropoutRNN}. Dropout masks are applied to the inputs, memory states and outputs of the RNN, and are also fixed across time steps. For memory updates, the RNN uses the Exponential Linear Unit (ELU) activation function.
\paragraph{Task-specific Layers}
For the task-specific layers, variables are grouped according to the sub-model types in Section \ref{sec:JointModels}, with layer $\mathbf{z_{c,t}}$ for continuous-valued longitudinal variables, $\mathbf{z_{b,t}}$ for binary longitudinal variables and $\mathbf{z_{e,t}}$ for events. Dropout masks are also applied to the outputs of each layer here. At the inputs to the continuous and binary task layers, a prediction horizon $\tau$ is also concatenated with the outputs from the RNN. This allows the parameters of the predictive distributions at $t+\tau$ to be computed in the final layer, i.e. $\mathbf{ \tilde{h}_{t}} = \left[ \mathbf{h}_t , \tau  \right]$.
\begin{subequations}
\begin{align}
\mathbf{z_{c,t}} = \text{ELU}( \mathbf{W_c} \mathbf{\tilde{h}_t} +  \mathbf{a_c})\\
\mathbf{z_{b,t}} = \text{ELU}( \mathbf{W_b} \mathbf{\tilde{h}_t} +  \mathbf{a_b})\\
\mathbf{z_{e,t}} = \text{ELU}( \mathbf{W_e} \mathbf{h_t} +  \mathbf{a_e})
\end{align}
\end{subequations}
%
%
\paragraph{Output Layer}
The final layer computes the parameter vectors of the predictive distribution, which are used to compute log likelihoods during training and dynamic predictions at run-time.
\begin{subequations}
\begin{align}
\mathbf{\mu_{t+\tau}} &=  \mathbf{W_{\mu}} \mathbf{z_{c,t}} +  \mathbf{a_{\mu}}  \\
\mathbf{\sigma_{t+\tau}}& = \text{Softplus}( \mathbf{W_{\sigma}} \mathbf{ z_{c,t}} +  \mathbf{a_{\sigma}})) \\
\mathbf{p_{t+\tau}}& = \text{Sigmoid}( \mathbf{W_{p}} \mathbf{ z_{b,t}} +  \mathbf{a_{p}})) \\
\lambda_{t} & = \text{Softplus}( \mathbf{W_{\lambda}} \mathbf{ z_{e,t}} +  \mathbf{a_{\lambda}}))  
\end{align}
\end{subequations}
Softplus activation functions are applied to $\mathbf{\sigma_{t+\tau}}$ and $\mathbf{p_{t+\tau}}$ to ensure that we obtain valid (i.e. $\ge 0$) standard deviations and binary probabilities. For simplicity, the exponential distribution is selected to model survival times, and predictive distributions can be expressed in a similar manner to that of Section \ref{sec:JointModels}:
\begin{subequations}
\label{eqn:submodeldist}
\begin{align}
Y_{t+\tau}^{(c)} &\sim N\left(\mu_{t+\tau}^{(c)}, \sigma_{t+\tau}^{ (c)  2}\right)\\
B_{t+\tau}^{(d)} &\sim \text{Bernoulli} \left(p_{t+\tau} ^{(d)} \right)\\
T_t^{(m)} &\sim \text{Exponential} \left( \lambda_{t}^{(m)} \right)
\end{align}
\end{subequations}
\subsection{Multitask Learning}
\label{sec:MultitaskLearning}
From the above, the negative log-likelihood of the data given the network is:
\begin{align}
\label{eqn:multivariateloss}
\mathcal{L}(\mathbf{W}) = 
\sum_{i,t,w,k_c,k_b,m} - \biggl[& \log f_c\left(Y_{i,t+\tau}^{(c)} |\mu_{t+\tau}^{(c)}, \sigma_{t+\tau}^{(c) 2},  \mathbf{W} \right) 
+ \log f_b\left(B_{i,t+\tau}^{(d)} |p_{t+\tau} ^{(d)}, \mathbf{W} \right) & \nonumber\\
& + \log f_T \left(T_{i,t}^{(m)} | \lambda_{t}^{(m)}, \mathbf{W} \right)\biggr]&
\end{align} 
Where $f_c(.), f_b(.)$ are likelihood functions based on Equations \ref{eqn:submodeldist} and $\mathbf{W}$ collectively represents the weights and biases of the entire network. For survival times, $f_T(.)$ is given as:
\begin{equation}
f_T \left(T_t^{(m)} | \lambda_{t}^{(m)}, \mathbf{W} \right)  = \left(\lambda_{t}^{(m)} \right)^{\delta_{i,T}} \exp\left(-\lambda_{t}^{(m)} T_t^{(m)}\right)
\end{equation} 
Which corresponds to event-free survival until time T before encountering the event \citep{generalizedLinearModels}. While the negative log-likelihood can be directly optimized across tasks, the use of multitask learning can yield the following benefits:
\paragraph{Better Survival Representations} As shown in \citep{ADMultitaskClassification}, multitask learning problems which have one main task of interest can weight the individual loss contributions of each subtask to favor representations for the main problem. For our current architecture, where we group similar tasks into task-specific layers, our loss function corresponds to:
\begin{align}
\label{eqn:multioutputloss}
L(\mathbf{W}) = &- \underbrace{\alpha_c \sum^{i,t,w,c} \log f_c\left(Y_{t+\tau}^{(c)} | \mathbf{W} \right) }_{\text{Continuous Longitudinal Loss } l_c } 
- \underbrace{\alpha_b \sum^{i,t,w,d}  \log f_b\left(B_{t+\tau}^{(d)} | \mathbf{W} \right)}_{\text{Binary Longitudinal Loss } l_b} & \nonumber\\
 &- \underbrace{\alpha_T \sum^{i,t,m} \log f_T \left(T_{t}^{(m)} | \mathbf{W} \right)}_{\text{Time-to-event Loss } l_T } &
\end{align} 
Given that survival predictions are the primary focus of many longitudinal studies, we set $\alpha_c = \alpha_b = 1$ and include $\alpha_T$ as an additional hyperparameter to be optimized. To train the network, patient trajectories are subdivided into Q sets of $\Omega_q(i, \rho, \tau) = \left\lbrace \mathbf{X_{i,{0:\rho}}}, \mathbf{Y_{i, \rho+\tau}}, \mathbf{T_{max,i}}, \mathbf{\delta_i} \right\rbrace$, where $\rho$ is the length of the covariate history to use in training trajectories up to a maximum of $\rho_{\max}$. Our procedure follows that of \citep{Collobert2008Multitasking}, as detailed in Algorithm \ref{alg:multitasktraining}.
\paragraph{Handling Irregularly Sampled Data} We address issues with irregular sampling by grouping variables that are measured together into the same task, and training the network with multitask learning. For instance, height, weight and BMI measurements are usually taken at the same time during follow-up, and can be grouped together in the same task. Given the completeness of the datasets we consider, we assume that task groupings match those defined by the task-specific layer of the network, and multitask learning is performed using Equation \ref{eqn:multioutputloss} and Algorithm \ref{alg:multitasktraining}. 

We note, however, that in the extreme case where none of the trajectories are aligned, we can define each variable as a separate task with its own loss function $l_*$. Algorithm \ref{alg:multitasktraining} then samples loss functions for one variable at a time, and the network is trained using only actual observations as target labels. This could reduce errors in cases where multiple sample rates exist and simple imputation is used, which might result in the multioutput networks replicating the imputation process instead of making true predictions. 

\begin{algorithm}[tbp]
   \caption{Training Disease-Atlas}
   \label{alg:multitasktraining}
\begin{algorithmic}
   \STATE {\bfseries Input:} Data $\Omega=\{\Omega_1, \dots, \Omega_Q \}$, max iterations $\mathcal{J}$
   \STATE {\bfseries Output:} Calibrated network weights $\mathbf{W}$
   \FOR{ \texttt{count}$=1$ {\bfseries to} $\mathcal{J}$}
   \STATE Get minibatch $\mathcal{M} \sim$  $\gamma$ random samples from $\Omega$
   \STATE Sample task loss function $l \sim \{l_c, l_b, l_T\}$ 
   \STATE Update $\mathbf{W} \leftarrow \texttt{Adam}(l, \mathcal{M})$, using feed-forward passes with  dropout applied
   \ENDFOR
\end{algorithmic}
\end{algorithm}
\subsection{Forecasting Disease Trajectories}
\label{sec:dropout}
Dynamic prediction involves 2 key elements - 1) calculating the expected longitudinal values and survival curves as described in Section \ref{sec:DynamicPredictionDefinition}, and 2) computing uncertainty estimates. To obtain these measures, we apply the Monte-Carlo dropout approach of \citep{GalDropoutRNN} by approximating the posterior over network weights as:
\begin{equation}
p(V^{(k)}_{t+\tau} | \mathcal{F}_t) \approx \frac{1}{J} \sum_{j=1}^J p(V^{(k)}_{t+\tau} | \mathcal{F}_t, \hat{\mathbf{W}}_j )
\end{equation}
Where we draw  $J$ samples $\hat{\mathbf{W}}_j$ using feed-forward passes through the network with the same dropout mask applied across time-steps. The samples obtained can then be used to compute expectations and uncertainty intervals for forecasts. 



\section{Tests on Medical Data}
\label{sec:Experiment}
\subsection{Overview of Datasets}
The UK Cystic Fibrosis (CF) registry contains data obtained for a cohort of 10980 CF patients during annual follow ups between 2008-2015, with a total of 87 variables associated with each patient across all years. In our investigations below, we consider a joint model for 2 continuous lung function scores (FEV1 and Predicted FEV1), 20 comorbidity and infection risks (treated as binary longitudinal observations) as well as death as the event of interest, simultaneously forecasting them all at each time step. We refer the reader to Appendix A for a full breakdown of the dataset. 

\subsection{Benchmarks and Training Procedure}
We compared the Disease-Atlas (DA) against simpler neural networks i.e. LSTM and Multi-layer Perceptrons (MLP), and traditional dynamic prediction methods, i.e. landmarking (L) \citep{vanHouwelingenDynamicPrediction} and joint models (JM). The data was partitioned into 3 sets: a training set with 60\% of the patients, a validation set with 20\% and a testing set with the final 20\%. Hyper-parameter optimization was performed using 20 iterations of random search on the validation data, and the test set was reserved for out-of-sample testing. Full details on the training procedure can be found in Appendix A.  Models were compared using several metrics: Mean-Squared Error (MSE) for predictions of continuous longitudinal variables, and the area under the Receiver Operating Characteristic (AUROC) and Precision-Recall Curve (AUPRC) for binary variables and the event of interest, i.e. death. Predictions were made via the MC dropout procedure described in Section \ref{sec:dropout}, with expectations computed using 300 samples per time step.

\paragraph{Disease-Atlas}  
Through subsequent tests, we demonstrate the performance contributions of the different innovations of Disease-Atlas, namely the usage of multitask learning in the presence of irregular sampling (Section \ref{sec:MultitaskMissingness}), and the inclusion of the RNN in the base layer for temporal information (Section \ref{sec:benchmarktests}). For the temporal layer, we evaluate the use of both a LSTM (DA-LSTM) or a single ELU layer (DA-NN).

\paragraph{Standard Neural Networks}  
Both the LSTM and MLP were taken to be benchmarks, structured to produce the same output distribution parameters as the Disease-Atlas, and optimized according to the multioutput loss function in Equation \ref{eqn:multivariateloss}. This makes the benchmarks equivalent to the DA-LSTM and DA-NN structures without task-specific layers and input $\tau$, and restricts them to making one-step-ahead longitudinal predictions only. 

\paragraph{Landmarking} For consistency with other studies of Cystic Fibrosis \citep{CF}, we use age as the time variable, and fit separate Cox regression models for patients in different age groups ($<25$, $25-50$, $50-75$ and $>75$ years old). As data is left-truncated with respect to age, we use the entry-exit implementation of the Cox proportional hazards model implemented in \citep{survival-package}. To avoid issues with collinearity, we start with a preliminary feature selection step first - performing multi-step Cox regression on the validation dataset, and only retaining features with coefficient p-values $<0.1$.

\paragraph{Joint Models} With the computational limitations of standard joint models, direct application to our dataset - containing over 6,500 patients in the training set with 87 annual covariate measurements - proved to be infeasible. As such, we used the two-step estimation procedure of \citep{MixedEffectsModelsForComplexData}, fitting the individual linear mixed effects (LME) longitudinal sub-models first, and using the mean estimates in the Cox regression model. For continuous variables, the linear mixed effects models used random intercepts and slopes, this is in line with the FEV1 models in \citep{FEV1a}, \citep{FEV1b} and \citep{FEV1c}. For binary variables, we use the logit regression version of the LME model, as implemented in the \texttt{lme4} R package \citep{lme4}.

\subsection{Evaluating Multitask Learning with Irregularly Sampled Data}
\label{sec:MultitaskMissingness}
To evaluate the effectiveness of multitask learning, we simulate irregular sampling by randomly removing all data points across each task (defined in Section  \ref{sec:MultitaskLearning}) with a probability $\gamma$ at every time step. For multivariate prediction, continuous-valued inputs were imputed using the mean value of the training set, while binary variables and indicators of death were set to 0. The networks were trained according to Section  \ref{sec:MultitaskLearning}, and then evaluated on the complete test set based on 1) MSE for continuous variables, and 2) AUROC for binary/event predictions.

\begin{figure*}[hptb]
\centering
\subfigure[FEV1 MSE]{
\includegraphics[width=0.5\defaultfigsize]{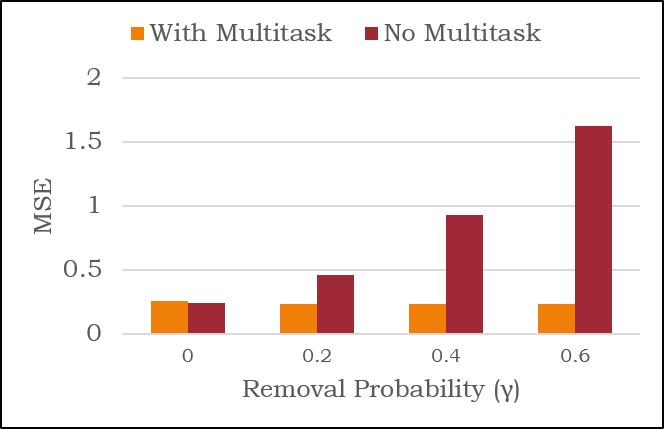}
}
\subfigure[Comorbidity AUROC]{
\includegraphics[width=0.5\defaultfigsize]{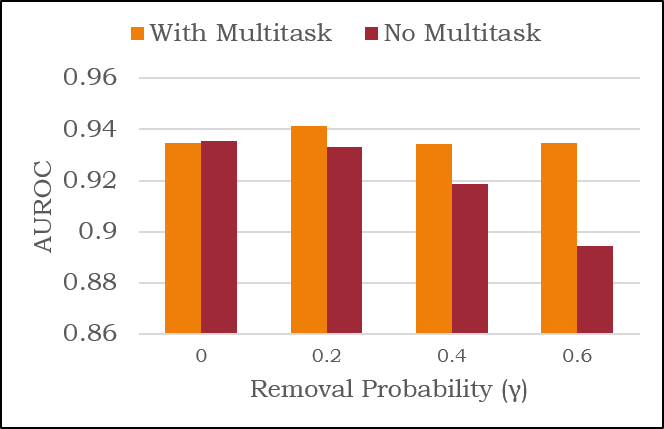}
}
\subfigure[Mortality AUROC]{
\includegraphics[width=0.5\defaultfigsize]{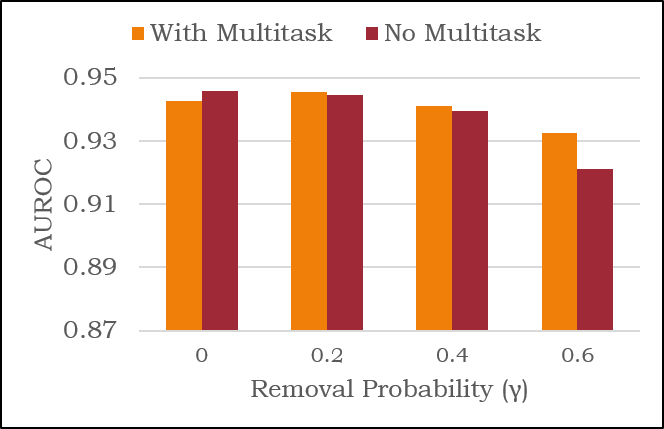}
}
\caption{Performance Comparison Between Multitask and Multioutput Networks}
\label {fig:multitask}
\end{figure*}

Figure \ref{fig:multitask} shows the outperformance of multitask learning, which has a lower MSE for FEV1 and higher AUROCs for comorbidity and mortality predictions as $\gamma$ increases. The improvements are most pronounced for MSE, as FEV1 has values at every time step, as opposed to binary observations and occurrences of death which are relatively more infrequent in the dataset. This demonstrates the robustness of the model to irregular sampling, and provides a way for joint models to be used on datasets even under such conditions.
\subsection{Performance vs Benchmarks}
\label{sec:benchmarktests}
As survival analysis is the usually main task of interest, we perform a comprehensive evaluation across all benchmarks, computing a probability of the event of interest, i.e. death, at a given time step using $1 - S_i^{(m)}(\tau| t)$ from Equation \ref{eqn:survivaldefn}. Performance was compared on the basis of AUROC and AUPRC of mortality predictions at various horizons $\tau$. To account for the stochastic nature of the MC dropout sampling procedure, performance evaluation was repeated 3 times for each neural network, with the averages and standard deviations of the metrics in reported in Table \ref{tab:SurvivalAuc}. 
\begin{table*}[h]
\centering
\caption{Results of Mortality Predictions for Cystic Fibrosis (Mean $\pm$ S.D. Across 3 Runs) }
\label{tab:SurvivalAuc}
\begin{tabular}{@{}l|l|llll|ll@{}}
\toprule
               & \textbf{$\tau$} & \textbf{DA-LSTM}              & \textbf{DA-NN}       & \textbf{LSTM}        & \textbf{MLP}                  & \textbf{L} & \textbf{JM} \\ \midrule
\textbf{AUROC} & 1               & \textbf{0.944($\pm$ 0.0004)}  & 0.943($\pm$ 0.0003)  & 0.943($\pm$ 0.0007)  & 0.941($\pm$ 0.0003)           & 0.824                & 0.870       \\
               & 2               & \textbf{0.924($\pm$ 0.0008)}  & 0.923($\pm$ 0.0005)  & 0.923($\pm$ 0.0005)  & 0.919($\pm$ 0.0003)           & 0.812                & 0.870       \\
               & 3               & \textbf{0.910($\pm$ 0.0003)}  & 0.905($\pm$ 0.0002)  & 0.908($\pm$ 0.0002)  & 0.907($\pm$ 0.0002)           & 0.825                & 0.851       \\
               & 4               & \textbf{0.905($\pm$ 0.0003)}  & 0.902($\pm$ 0.0008)  & 0.904($\pm$ 0.0003)  & 0.904($\pm$ 0.0006)           & 0.776                & 0.828       \\
               & 5               & \textbf{0.895($\pm$ 0.0003)}  & 0.892($\pm$ 0.0005)  & 0.894($\pm$ 0.0005)  & 0.888($\pm$ 0.0007)           & 0.765                & 0.806       \\ \midrule
\textbf{AUPRC} & 1               & \textbf{0.278 ($\pm$ 0.0037)} & 0.238 ($\pm$ 0.0040) & 0.230 ($\pm$ 0.0020) & 0.219 ($\pm$ 0.0036)          & 0.161                & 0.119       \\
               & 2               & \textbf{0.193 ($\pm$ 0.0014)} & 0.169 ($\pm$ 0.0033) & 0.165 ($\pm$ 0.0017) & 0.186 ($\pm$ 0.0036)          & 0.082                & 0.092       \\
               & 3               & 0.103 ($\pm$ 0.0005)          & 0.092 ($\pm$ 0.0007) & 0.099 ($\pm$ 0.0028) & \textbf{0.105 ($\pm$ 0.0001)} & 0.085                & 0.089       \\
               & 4               & \textbf{0.109 ($\pm$ 0.0007)} & 0.101 ($\pm$ 0.0014) & 0.095 ($\pm$ 0.0010) & 0.102 ($\pm$ 0.0006)          & 0.062                & 0.068       \\
               & 5               & \textbf{0.101 ($\pm$ 0.0007)} & 0.091 ($\pm$ 0.0008) & 0.093 ($\pm$ 0.0017) & 0.100 ($\pm$ 0.0017)          & 0.058                & 0.059       \\ \bottomrule
\end{tabular}
\end{table*}

The first thing to note is the vast improvements of neural networks over traditional dynamic prediction models, with the DA-LSTM showing average AUPRC improvements of $75\%$ and $78\%$ across all time steps when compared to landmarking and standard joint models. Among the neural network benchmarks, the strongest gains for the DA-LSTM can be seen over shorter prediction horizons (1-2 years), improving the DA-NN by $17\%$ and the LSTM by $21\%$ in one-step-ahead AUPRC. This highlights the benefits of both the use of temporal information and the addition of task-specific layers for short-term survival predictions. While the gains in AUROC are indeed smaller, this can be attributed to the large class imbalance present within the dataset, due to the rare occurrence of death in the dataset (seen in 451 of 10275 patients) and the censoring effect it has on a patients trajectory. As shown in \citep{ClassImbalance}, ROC metrics can lead to deceptive good performance, as the definition of the false positive rate (false positive / total number of negatives) permits the occurrence of a large number of false positives in imbalanced datasets, and recommend the use of PRC metrics. In addition, we note that the Disease-Atlas architectures also allow for the forecasting of longitudinal variables over arbitrary horizons, which standard neural network architectures are unable to accommodate by default.
%
\begin{table*}[hptb]
\centering
\caption{Results of Longitudinal Predictions for Cystic Fibrosis (Single Run)}
\label{tab:LongitudinalResults}
\centerline{
\begin{tabular}{@{}l|l|ll|ll|ll@{}}
\toprule
                 & \textbf{} & \multicolumn{2}{c|}{\textbf{MSE}} & \multicolumn{2}{c|}{\textbf{AUROC {[}Mean $\pm$ SD{]}}}       & \multicolumn{2}{c}{\textbf{AUPRC {[}Mean $\pm$SD{]}}}         \\
                 & $\tau$    & FEV1            & Pred. FEV1      & Comorbidities               & Infections                  & Comorbidities               & Infections                  \\ \midrule
\textbf{DA-LSTM} & 1         & \textbf{0.182}  & \textbf{121.3}  & \textbf{0.957 ($\pm$ 0.025)} & \textbf{0.888 ($\pm$ 0.056)} & \textbf{0.680 ($\pm$0.261)}  & \textbf{0.416 ($\pm$0.247)}  \\
                 & 2         & \textbf{0.191}  & \textbf{139.4}  & \textbf{0.926 ($\pm$0.047)}  & \textbf{0.850 ($\pm$ 0.044)} & \textbf{0.648 ($\pm$ 0.244)} & \textbf{0.337 ($\pm$ 0.261)} \\
                 & 3         & \textbf{0.275}  & \textbf{191.3}  & \textbf{0.882 ($\pm$0.048)}  & \textbf{0.798 ($\pm$ 0.057)} & \textbf{0.555 ($\pm$ 0.213)} & \textbf{0.337 ($\pm$ 0.261)} \\
                 & 4         & \textbf{0.374}  & \textbf{254.4}  & \textbf{0.817 ($\pm$0.085)}  & \textbf{0.723 ($\pm$ 0.068)} & \textbf{0.459 ($\pm$ 0.184)} & \textbf{0.309 ($\pm$ 0.252)} \\
                 & 5         & \textbf{0.461}  & \textbf{308.1}  & \textbf{0.790 ($\pm$0.067)}  & \textbf{0.669 ($\pm$ 0.126)} & \textbf{0.388 ($\pm$ 0.169)} & \textbf{0.269 ($\pm$ 0.247)} \\ \midrule
\textbf{JM}      & 1         & 0.553           & 368.6           & 0.699 ($\pm$ 0.148)             & 0.673 ($\pm$ 0.069)             & 0.176 ($\pm$ 0.088)             & 0.161 ($\pm$ 0.176)             \\
                 & 2         & 0.593           & 411.1           & 0.694 ($\pm$ 0.139)             & 0.651 ( $\pm$ 0.060)            & 0.180 ( $\pm$ 0.089)            & 0.157 ( $\pm$ 0.181)            \\
                 & 3         & 0.641           & 451.8           & 0.685 ($\pm$ 0.140)             & 0.631 ( $\pm$ 0.072)            & 0.185 ( $\pm$ 0.090)            & 0.160 ( $\pm$ 0.186)            \\
                 & 4         & 0.695           & 490.1           & 0.681 ($\pm$ 0.132)             & 0.607 ( $\pm$ 0.077)            & 0.187 ( $\pm$ 0.091)            & 0.159 ( $\pm$ 0.188)            \\
                 & 5         & 0.750           & 519.7           & 0.673 ($\pm$ 0.130)             & 0.580 ( $\pm$ 0.082)            & 0.188 ( $\pm$ 0.093)            & 0.155 ( $\pm$ 0.186)            \\ \bottomrule
\end{tabular}}
\end{table*}

We proceed to evaluate the longitudinal forecasting ability of the Disease-Atlas in Table \ref{tab:LongitudinalResults}, focusing on comparisons between the DA-LSTM and standard joint models. To provide a concise summary of longitudinal prediction results, we use a single set of 300 MC dropout samples to compute performance metrics. The average AUROC/AUPRC for comorbidities and infections are reported separately, along with standard deviations across each group. Similarly, results for FEV1 and Predicted FEV1 are reported for a single evaluation, and a full breakdown of results for each individual binary longitudinal variable is detailed in Appendix A. We can see that the DA-LSTM improves performance across all longitudinal prediction categories, reducing MSEs by $56\%$ on average across all continuous predictions, and improving AUPRCs in comorbidities and infections on average by $199\%$ and $112\%$  respectively. This vast improvement underscores the ability of deep neural networks to learn complex interactions directly from the data, and demonstrates the benefits of using a deep learning approach to joint modeling.

\section{Illustrative Use Case for Disease-Atlas: Personalized Screening}
While numerous use cases for the Disease-Atlas exist, one possible example is its use in personalized screening, i.e. prescribing testing regimes and follow-up schedules that are tailored to the unique characteristics of a patient. \citep{DPScreen} derive an optimal policy that balances the costs of screening, along with the risks of delaying the screening process. Their paper, however, has two important limitations: 1) it requires the evolution of the disease to be known, and 2) it requires an analytical expression for the cost of delay, which is often difficult to determine in practice. Disease-Atlas does not suffer from these limitations. 

We illustrate how the Disease Atlas can be used for identifying screening profiles for Cystic Fibrosis patients. We use as an exemplar an actual patient from our test set who started to be seen in 2008 and is screened for Diabetes, a very important comorbidity affecting the treatment of patients.

\begin{figure*}[h]
\centerline{\includegraphics[width=2\defaultfigsize]{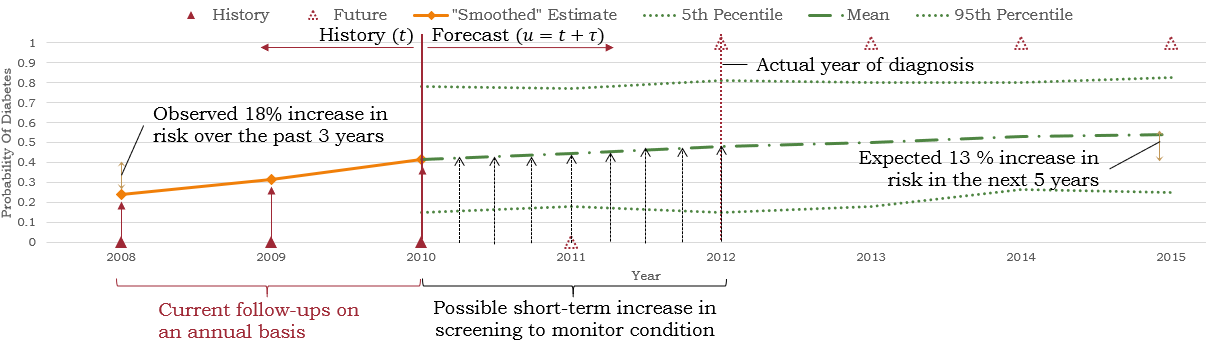}}
\caption{Using Disease-Atlas for Personalized Screening}
\label {fig:personalised_screening}
\end{figure*}

Figure \ref{fig:personalised_screening} shows how the Disease Atlas can be used to design personalized screening policies. Using the Disease Atlas as applied from 2008-2010, we can record the ``smoothed'' estimate at each time step, i.e. $\hat{B}_i^{(d)} (0|t)$ as per Equation \ref{eqn:DynPredExpectation} and plotted in orange. To determine a patient's future risk, we extrapolate in the usual way over various horizons $\tau$, providing both the expected value and an uncertainty interval using the 5th and 95th percentiles of the MC dropout samples. From Figure 6, we see that the patient had a steady 18\%
increase in risk from 2008 to 2010, and will expect a further increase of 13\% over the next 5 years. Informed by the Disease Atlas, the clinician may decide to prescribe additional tests for Diabetes, or increase the frequency of follow-up in the short-term to better monitor risks.

\newpage
\bibliography{disease_atlas}
\newpage
\appendix
\part*{Appendix for Disease-Atlas}
\section{Tests on Cystic Fibrosis Dataset}
\subsection{Details on Dataset}
The UK Cystic Fibrosis (CF) registry contains data obtained for a cohort of 10980 CF patients during annual follow ups between 2008-2015, with a total of 87 variables that were associated with each patient across all years. This includes demographic information (e.g. age, height, weight, BMI), genetic information, treatments received, metrics for lung function (FEV1 and Predicted FEV1), comorbidities observed, and any bacterial infections developed. In our investigations, we consider a joint model for the 2 continuous lung function scores (FEV1 and Predicted FEV1), 20 binary longitudinal variables of comorbidity and infection, along with death as the event of interest. 

A full description of the jointly-modeled longitudinal and time-to-event datasets can be found in Table \ref{tab:cfstats}.

\begin{table*}[b]
\centering
\caption{Description of Longitudinal and Time-to-event Data for CF}
\label{tab:cfstats}
\begin{tabular}{@{}llllllll@{}}
\toprule
                       & \textbf{}              & \textbf{Type}  & \textbf{\% Patients} & \textbf{Mean} & \textbf{S.D.} & \textbf{Min} & \textbf{Max} \\ \midrule
\textbf{Event}         & Death                  & Binary (Event) & 4.70\%                              & 0.008         & 0.087                       & 0.000        & 1.000        \\ \midrule
\textbf{Biomarkers}    & FEV1                   & Continuous     & 100.00\%                            & 2.176         & 0.914                       & 0.090        & 6.250        \\
\textbf{}              & Predicted FEV1         & Continuous     & 100.00\%                            & 72.109        & 22.404                      & 8.950        & 197    \\ \midrule
\textbf{Comorbidities} & Liver Disease          & Binary         & 20.80\%                             & 0.128         & 0.334                       & 0.000        & 1.000        \\
                       & Asthma                 & Binary         & 22.96\%                             & 0.146         & 0.353                       & 0.000        & 1.000        \\
                       & Arthropathy            & Binary         & 9.50\%                              & 0.050         & 0.218                       & 0.000        & 1.000        \\
                       & Bone fracture          & Binary         & 1.94\%                              & 0.007         & 0.081                       & 0.000        & 1.000        \\
                       & Raised Liver Enzymes   & Binary         & 23.91\%                             & 0.114         & 0.318                       & 0.000        & 1.000        \\
                       & Osteopenia             & Binary         & 20.37\%                             & 0.114         & 0.318                       & 0.000        & 1.000        \\
                       & Osteoporosis           & Binary         & 9.58\%                              & 0.051         & 0.219                       & 0.000        & 1.000        \\
                       & Hypertension           & Binary         & 3.30\%                              & 0.020         & 0.139                       & 0.000        & 1.000        \\
                       & Diabetes               & Binary         & 24.56\%                             & 0.167         & 0.373                       & 0.000        & 1.000        \\ \midrule
\textbf{Bacterial}     & Burkholderia Cepacia   & Binary         & 5.59\%                              & 0.034         & 0.181                       & 0.000        & 1.000        \\
\textbf{Infections}    & Pseudomonas Aeruginosa & Binary         & 65.18\%                             & 0.407         & 0.491                       & 0.000        & 1.000        \\
                       & Haemophilus Influenza  & Binary         & 30.55\%                             & 0.091         & 0.288                       & 0.000        & 1.000        \\
                       & Aspergillus            & Binary         & 29.29\%                             & 0.110         & 0.313                       & 0.000        & 1.000        \\
                       & NTM                    & Binary         & 6.38\%                              & 0.019         & 0.136                       & 0.000        & 1.000        \\
                       & Ecoli                  & Binary         & 5.32\%                              & 0.012         & 0.111                       & 0.000        & 1.000        \\
                       & Klebsiella Pneumoniae  & Binary         & 4.93\%                              & 0.010         & 0.101                       & 0.000        & 1.000        \\
                       & Gram-Negative          & Binary         & 3.78\%                              & 0.008         & 0.089                       & 0.000        & 1.000        \\
                       & Xanthomonas            & Binary         & 13.18\%                             & 0.043         & 0.202                       & 0.000        & 1.000        \\
                       & Staphylococcus Aureus  & Binary         & 52.59\%                             & 0.244         & 0.429                       & 0.000        & 1.000        \\
                       & ALCA                   & Binary         & 5.06\%                              & 0.020         & 0.138                       & 0.000        & 1.000        \\ \bottomrule
\end{tabular}
\end{table*}

\subsection{Hyperparameter Optimization}
\label{sec:hyperparamopt}

Hyperparameter optimization was conducted using 20 iterations of random search, with the search space documented in Table \ref{tab:paramoptions}. Please note that the RNN state size was defined relative to the number of input features ($L$). The task specific layers were also size in relation to the RNN state size, and defined to be (state size + task output size) /2. This was done to ensure that we had a principled way of sizing the task-specific layer relative to state size and outputs, without having to add on too many additional hyper-parameters (i.e. one per task-specific layer). In addition, $\alpha_T$ was defined in relation to the number of longitudinal variables ($K$). All neural networks were trained to convergence, as determined by the survival log-likelihoods evaluated on the validation data, or up to a maximum of 50 epochs.

The final parameters obtained for each network can be found in Table \ref{tab:paramselected}.

\begin{table*}[tp]
\centering
\caption{Hyper-parameter Selection Range for Random Search}
\label{tab:paramoptions}
\begin{tabular}{@{}ll@{}}
\toprule
                               & \textbf{Hyper-parameter Selection Range} \\ \midrule
\textbf{Max Number of  Epochs} & 50                                       \\
\textbf{RNN State Size}       & 1L, 2L, 3L, 4L, 5L            \\
\textbf{$\alpha_T$}            & K, 2K, 3K, 4K, 5K                        \\
\textbf{Max Gradient Norm}              & 0.5, 1.0, 1.5, 2.0                       \\
\textbf{Learning Rate}         &  1e-3, 5e-3, 1e-4                 \\
\textbf{Minibatch Size}        & 64, 128, 256                          \\
\textbf{Dropout Rate}          & 0.2, 0.3, 0.4, 0.5 \\ \bottomrule
\end{tabular}
\end{table*}

\begin{table*}[]
\centering
\caption{Hyper-parameters Selected for CF Tests}
\label{tab:paramselected}
\centerline{
\begin{tabular}{@{}lllllll@{}}
\toprule
                               & \textbf{State Size} & \textbf{Minibatch Size} & \textbf{Learning Rate} & \textbf{Max Gradient Norm} & \textbf{Dropout Rate} & \textbf{$\alpha_T$} \\ \midrule
\textbf{DA-LSTM}               & 1L          & 256                & 1.00E-04               & 0.5                        & 0.3                   & 3K          \\
\textbf{DA-NN}                 & 5L          & 256                & 1.00E-04               & 2.0                        & 0.3                   & 4K          \\
\textbf{Standard LSTM}         & 1L          & 32                 & 1.00E-03               & 1.5                        & 0.4                   & 4K          \\ 
\textbf{Standard NN}           & 1.5L        & 64                 & 1.00E-03               & 1                          & 0.45                  & 1K          \\
\bottomrule
\end{tabular}}
\end{table*}

\subsection{Additional Results}
To supplement the results in the test section of the main report, a detailed breakdown of the prediction results for binary longitudinal variables can be found in Table \ref{tab:CfAuroc} and \ref{tab:CfAuprc} for DA-LSTM and JM respectively. For the DA-LSTM, due to the randomness present in the MC dropout  procedure, the performance evaluation was repeated 3 times with the means and standard deviations reported in the tables. The results also demonstrate the outperformance of the deep neural network over standard benchmarks in both AUROC and AUPRC terms.

\begin{table*}[]
\centering
\caption{AUROC for Comorbidity and Infection Predictions for CF Dataset (Mean $\pm$ S.D. Across 3 Runs)}
\label{tab:CfAuroc}
\centerline{
\begin{tabular}{@{}llllll@{}}
\toprule
\textbf{DA-LSTM}             & \textbf{1}         & \textbf{2}         & \textbf{3}         & \textbf{4}         & \textbf{5}         \\ \midrule
\underline{\textbf{Comorbidities}} &                    &                    &                    &                    &                    \\
Liver Disease                & 0.975 ($\pm$ 0.0001) & 0.948 ($\pm$ 0.0001) & 0.897 ($\pm$ 0.0003) & 0.826 ($\pm$ 0.0005) & 0.767 ($\pm$ 0.0001) \\
Asthma                       & 0.979 ($\pm$ 0.0001) & 0.946 ($\pm$ 0.0002) & 0.906 ($\pm$ 0.0005) & 0.843 ($\pm$ 0.0004) & 0.784 ($\pm$ 0.0003) \\
Arthropathy                  & 0.975 ($\pm$ 0.0004) & 0.950 ($\pm$ 0.0002) & 0.911 ($\pm$ 0.0004) & 0.888 ($\pm$ 0.0007) & 0.833 ($\pm$ 0.0004) \\
Bone fracture                & 0.891 ($\pm$ 0.0024) & 0.791 ($\pm$ 0.0015) & 0.789 ($\pm$ 0.0014) & 0.610 ($\pm$ 0.0029) & 0.721 ($\pm$ 0.0013) \\
Raised Liver Enzymes         & 0.937 ($\pm$ 0.0006) & 0.909 ($\pm$ 0.0002) & 0.798 ($\pm$ 0.0004) & 0.741 ($\pm$ 0.0006) & 0.685 ($\pm$ 0.0002) \\
Osteopenia                   & 0.961 ($\pm$ 0.0005) & 0.942 ($\pm$ 0.0003) & 0.897 ($\pm$ 0.0006) & 0.873 ($\pm$ 0.0005) & 0.850 ($\pm$ 0.0004) \\
Osteoporosis                 & 0.956 ($\pm$ 0.0006) & 0.943 ($\pm$ 0.0008) & 0.912 ($\pm$ 0.0006) & 0.877 ($\pm$ 0.0008) & 0.854 ($\pm$ 0.0003) \\
Hypertension                 & 0.977 ($\pm$ 0.0004) & 0.955 ($\pm$ 0.0006) & 0.936 ($\pm$ 0.0009) & 0.879 ($\pm$ 0.0006) & 0.869 ($\pm$ 0.0009) \\
Diabetes                     & 0.963 ($\pm$ 0.0002) & 0.942 ($\pm$ 0.0002) & 0.918 ($\pm$ 0.0001) & 0.873 ($\pm$ 0.0003) & 0.844 ($\pm$ 0.0004) \\ \midrule
\underline{\textbf{Infections}}    &                    &                    &                    &                    &                    \\
Burkholderia Cepacia         & 0.960 ($\pm$ 0.0011) & 0.929 ($\pm$ 0.0009) & 0.922 ($\pm$ 0.0008) & 0.876 ($\pm$ 0.0004) & 0.846 ($\pm$ 0.0011) \\
Pseudomonas Aeruginosa       & 0.898 ($\pm$ 0.0004) & 0.878 ($\pm$ 0.0002) & 0.850 ($\pm$ 0.0002) & 0.818 ($\pm$ 0.0003) & 0.796 ($\pm$ 0.0005) \\
Haemophilus Influenza        & 0.848 ($\pm$ 0.0007) & 0.835 ($\pm$ 0.0002) & 0.798 ($\pm$ 0.0002) & 0.737 ($\pm$ 0.0007) & 0.738 ($\pm$ 0.0012) \\
Aspergillus                  & 0.873 ($\pm$ 0.0007) & 0.798 ($\pm$ 0.0004) & 0.789 ($\pm$ 0.0007) & 0.673 ($\pm$ 0.0010) & 0.665 ($\pm$ 0.0006) \\
NTM                          & 0.897 ($\pm$ 0.0008) & 0.802 ($\pm$ 0.0013) & 0.824 ($\pm$ 0.0014) & 0.675 ($\pm$ 0.0008) & 0.633 ($\pm$ 0.0002) \\
Ecoli                        & 0.929 ($\pm$ 0.0018) & 0.894 ($\pm$ 0.0013) & 0.733 ($\pm$ 0.0044) & 0.677 ($\pm$ 0.0043) & 0.340 ($\pm$ 0.0066) \\
Klebsiella Pneumoniae        & 0.950 ($\pm$ 0.0012) & 0.904 ($\pm$ 0.0007) & 0.815 ($\pm$ 0.0044) & 0.630 ($\pm$ 0.0005) & 0.725 ($\pm$ 0.0046) \\
Gram-Negative                & 0.745 ($\pm$ 0.0052) & 0.793 ($\pm$ 0.0045) & 0.690 ($\pm$ 0.0007) & 0.698 ($\pm$ 0.0020) & 0.587 ($\pm$ 0.0027) \\
Xanthomonas                  & 0.894 ($\pm$ 0.0009) & 0.831 ($\pm$ 0.0005) & 0.770 ($\pm$ 0.0013) & 0.716 ($\pm$ 0.0007) & 0.654 ($\pm$ 0.0006) \\
Staphylococcus Aureus        & 0.908 ($\pm$ 0.0002) & 0.856 ($\pm$ 0.0004) & 0.784 ($\pm$ 0.0005) & 0.699 ($\pm$ 0.0003) & 0.649 ($\pm$ 0.0005) \\
ALCA                         & 0.863 ($\pm$ 0.0008) & 0.831 ($\pm$ 0.0010) & 0.800 ($\pm$ 0.0008) & 0.758 ($\pm$ 0.0010) & 0.725 ($\pm$ 0.0015) \\ \midrule
\textbf{JM}                  & \textbf{1}         & \textbf{2}         & \textbf{3}         & \textbf{4}         & \textbf{5}         \\ \midrule
\underline{\textbf{Comorbidities}} &                    &                    &                    &                    &                    \\
Liver Disease                & 0.634              & 0.622              & 0.619              & 0.607              & 0.602              \\
Asthma                       & 0.701              & 0.671              & 0.649              & 0.622              & 0.597              \\
Arthropathy                  & 0.761              & 0.755              & 0.755              & 0.757              & 0.749              \\
Bone Fracture                & 0.344              & 0.379              & 0.378              & 0.413              & 0.411              \\
Raised Liver Enzymes         & 0.622              & 0.608              & 0.589              & 0.584              & 0.594              \\
Osteopenia                   & 0.774              & 0.767              & 0.761              & 0.758              & 0.746              \\
Osteoporosis                 & 0.801              & 0.794              & 0.781              & 0.772              & 0.76               \\
Hypertension                 & 0.894              & 0.891              & 0.893              & 0.893              & 0.889              \\
Diabetes                     & 0.76               & 0.754              & 0.739              & 0.723              & 0.709              \\ \midrule
\underline{\textbf{Infections}}    &                    &                    &                    &                    &                    \\
Burkholderia Cepacia         & 0.636              & 0.638              & 0.63               & 0.622              & 0.613              \\
Pseudomonas Aeruginosa       & 0.744              & 0.735              & 0.727              & 0.713              & 0.692              \\
Haemophilus Influenza        & 0.698              & 0.712              & 0.722              & 0.715              & 0.685              \\
Aspergillus                  & 0.716              & 0.689              & 0.662              & 0.622              & 0.603              \\
NTM                          & 0.709              & 0.686              & 0.678              & 0.633              & 0.586              \\
Ecoli                        & 0.729              & 0.604              & 0.507              & 0.444              & 0.389              \\
Klebsiella Pneumoniae        & 0.777              & 0.73               & 0.706              & 0.67               & 0.624              \\
Gram-Negative                & 0.533              & 0.55               & 0.53               & 0.518              & 0.489              \\
Xanthomonas                  & 0.628              & 0.613              & 0.604              & 0.611              & 0.603              \\
Staphylococcus Aureus        & 0.607              & 0.589              & 0.577              & 0.561              & 0.542              \\
ALCA                         & 0.624              & 0.613              & 0.598              & 0.572              & 0.552              \\ \bottomrule
\end{tabular}}
\end{table*}

\begin{table*}[]
\centering
\caption{AUPRC for Comorbidity and Infection Predictions for CF Dataset (Mean $\pm$ S.D. Across 3 Runs)}
\label{tab:CfAuprc}
\centerline{
\begin{tabular}{@{}llllll@{}}
\toprule
\textbf{DA-LSTM}             & \textbf{1}           & \textbf{2}           & \textbf{3}           & \textbf{4}           & \textbf{5}           \\ \midrule
\underline{ \textbf{Comorbidities}} &                      &                      &                      &                      &                      \\
Liver Disease                & 0.862 ($\pm$ 0.0006) & 0.825 ($\pm$ 0.0007) & 0.709 ($\pm$ 0.0022) & 0.616 ($\pm$ 0.0009) & 0.513 ($\pm$ 0.0012) \\
Asthma                       & 0.904 ($\pm$ 0.0006) & 0.845 ($\pm$ 0.0015) & 0.773 ($\pm$ 0.0016) & 0.642 ($\pm$ 0.0024) & 0.544 ($\pm$ 0.0019) \\
Arthropathy                  & 0.799 ($\pm$ 0.0036) & 0.760 ($\pm$ 0.0027) & 0.621 ($\pm$ 0.0011) & 0.500 ($\pm$ 0.0038) & 0.347 ($\pm$ 0.0026) \\
Bone fracture                & 0.064 ($\pm$ 0.0020) & 0.043 ($\pm$ 0.0012) & 0.052 ($\pm$ 0.0011) & 0.032 ($\pm$ 0.0011) & 0.031 ($\pm$ 0.0018) \\
Raised Liver Enzymes         & 0.784 ($\pm$ 0.0015) & 0.726 ($\pm$ 0.0019) & 0.536 ($\pm$ 0.0016) & 0.409 ($\pm$ 0.0024) & 0.338 ($\pm$ 0.0011) \\
Osteopenia                   & 0.758 ($\pm$ 0.0018) & 0.742 ($\pm$ 0.0013) & 0.648 ($\pm$ 0.0024) & 0.577 ($\pm$ 0.0009) & 0.526 ($\pm$ 0.0017) \\
Osteoporosis                 & 0.658 ($\pm$ 0.0044) & 0.644 ($\pm$ 0.0028) & 0.507 ($\pm$ 0.0017) & 0.406 ($\pm$ 0.0013) & 0.322 ($\pm$ 0.0011) \\
Hypertension                 & 0.308 ($\pm$ 0.0069) & 0.340 ($\pm$ 0.0072) & 0.309 ($\pm$ 0.0074) & 0.277 ($\pm$ 0.0031) & 0.227 ($\pm$ 0.0010) \\
Diabetes                     & 0.850 ($\pm$ 0.0006) & 0.798 ($\pm$ 0.0019) & 0.774 ($\pm$ 0.0013) & 0.663 ($\pm$ 0.0020) & 0.640 ($\pm$ 0.0034) \\ \midrule
\underline{ \textbf{Infections}}    &                      &                      &                      &                      &                      \\
Burkholderia Cepacia         & 0.692 ($\pm$ 0.0029) & 0.672 ($\pm$ 0.0026) & 0.639 ($\pm$ 0.0047) & 0.576 ($\pm$ 0.0058) & 0.471 ($\pm$ 0.0052) \\
Pseudomonas Aeruginosa       & 0.840 ($\pm$ 0.0002) & 0.828 ($\pm$ 0.0010) & 0.815 ($\pm$ 0.0010) & 0.800 ($\pm$ 0.0007) & 0.794 ($\pm$ 0.0017) \\
Haemophilus Influenza        & 0.369 ($\pm$ 0.0006) & 0.332 ($\pm$ 0.0005) & 0.265 ($\pm$ 0.0007) & 0.243 ($\pm$ 0.0009) & 0.278 ($\pm$ 0.0007) \\
Aspergillus                  & 0.380 ($\pm$ 0.0038) & 0.315 ($\pm$ 0.0008) & 0.337 ($\pm$ 0.0019) & 0.270 ($\pm$ 0.0011) & 0.293 ($\pm$ 0.0012) \\
NTM                          & 0.237 ($\pm$ 0.0008) & 0.073 ($\pm$ 0.0006) & 0.181 ($\pm$ 0.0017) & 0.133 ($\pm$ 0.0024) & 0.138 ($\pm$ 0.0008) \\
Ecoli                        & 0.506 ($\pm$ 0.0040) & 0.242 ($\pm$ 0.0030) & 0.089 ($\pm$ 0.0021) & 0.036 ($\pm$ 0.0005) & 0.008 ($\pm$ 0.0009) \\
Klebsiella Pneumoniae        & 0.299 ($\pm$ 0.0039) & 0.146 ($\pm$ 0.0044) & 0.060 ($\pm$ 0.0041) & 0.010 ($\pm$ 0.0000) & 0.015 ($\pm$ 0.0004) \\
Gram-Negative                & 0.028 ($\pm$ 0.0007) & 0.038 ($\pm$ 0.0013) & 0.022 ($\pm$ 0.0004) & 0.027 ($\pm$ 0.0003) & 0.022 ($\pm$ 0.0004) \\
Xanthomonas                  & 0.298 ($\pm$ 0.0068) & 0.202 ($\pm$ 0.0037) & 0.218 ($\pm$ 0.0020) & 0.180 ($\pm$ 0.0022) & 0.128 ($\pm$ 0.0019) \\
Staphylococcus Aureus        & 0.771 ($\pm$ 0.0010) & 0.706 ($\pm$ 0.0018) & 0.612 ($\pm$ 0.0014) & 0.537 ($\pm$ 0.0002) & 0.497 ($\pm$ 0.0006) \\
ALCA                         & 0.153 ($\pm$ 0.0011) & 0.148 ($\pm$ 0.0024) & 0.155 ($\pm$ 0.0040) & 0.144 ($\pm$ 0.0019) & 0.175 ($\pm$ 0.0025) \\ \midrule
\textbf{JM}                  & \textbf{1}           & \textbf{2}           & \textbf{3}           & \textbf{4}           & \textbf{5}           \\ \midrule
\underline{\textbf{Comorbidities}} &                      &                      &                      &                      &                      \\
Liver Disease                & 0.181                & 0.186                & 0.197                & 0.2                  & 0.207                \\
Asthma                       & 0.272                & 0.261                & 0.258                & 0.245                & 0.24                 \\
Arthropathy                  & 0.134                & 0.142                & 0.148                & 0.155                & 0.154                \\
Bone Fracture                & 0.006                & 0.007                & 0.007                & 0.009                & 0.01                 \\
Raised Liver Enzymes         & 0.163                & 0.16                 & 0.156                & 0.157                & 0.172                \\
Osteopenia                   & 0.245                & 0.255                & 0.266                & 0.278                & 0.28                 \\
Osteoporosis                 & 0.144                & 0.149                & 0.151                & 0.146                & 0.134                \\
Hypertension                 & 0.123                & 0.13                 & 0.141                & 0.142                & 0.142                \\
Diabetes                     & 0.319                & 0.334                & 0.342                & 0.348                & 0.356                \\ \midrule
\underline{\textbf{Infections}}    &                      &                      &                      &                      &                      \\
Burkholderia Cepacia         & 0.054                & 0.058                & 0.056                & 0.056                & 0.062                \\
Pseudomonas Aeruginosa       & 0.636                & 0.641                & 0.65                 & 0.655                & 0.649                \\
Haemophilus Influenza        & 0.181                & 0.204                & 0.233                & 0.231                & 0.202                \\
Aspergillus                  & 0.22                 & 0.22                 & 0.218                & 0.212                & 0.216                \\
NTM                          & 0.076                & 0.068                & 0.072                & 0.062                & 0.041                \\
Ecoli                        & 0.098                & 0.037                & 0.025                & 0.011                & 0.005                \\
Klebsiella Pneumoniae        & 0.051                & 0.037                & 0.026                & 0.025                & 0.027                \\
Gram-Negative                & 0.009                & 0.01                 & 0.012                & 0.012                & 0.015                \\
Xanthomonas                  & 0.079                & 0.079                & 0.087                & 0.092                & 0.098                \\
Staphylococcus Aureus        & 0.336                & 0.337                & 0.344                & 0.347                & 0.345                \\
ALCA                         & 0.037                & 0.04                 & 0.037                & 0.04                 & 0.047                \\ \bottomrule
\end{tabular}}
\end{table*}

\section{Acknowledgments}
This work is supported by the UK Cystic Fibrosis Trust and the Oxford-Man Institute. We would also like to thank the UK Cystic Fibrosis Trust for providing us access to the data from the CF registry, which was used extensively in the analysis performed in this report.
\end{document}